\documentclass[conference]{IEEEtran}
\IEEEoverridecommandlockouts

\usepackage{cite}
\usepackage{amsmath,amssymb,amsfonts}
\usepackage{graphicx}
\usepackage{textcomp}
\usepackage{xcolor}
\usepackage{multirow}
\usepackage{makecell}
\usepackage{algorithm}
\usepackage{array}
\usepackage[caption=false,font=normalsize,labelfont=sf,textfont=sf]{subfig}
\usepackage{stfloats}
\usepackage{url}
\usepackage{verbatim}
\usepackage{booktabs}
\usepackage{cite}
\usepackage{hyperref}
\usepackage{tabularx}
\usepackage{algpseudocode}
\usepackage{adjustbox}
\usepackage{colortbl}
\usepackage{stmaryrd}
\def\BibTeX{{\rm B\kern-.05em{\sc i\kern-.025em b}\kern-.08em
    T\kern-.1667em\lower.7ex\hbox{E}\kern-.125emX}}
\begin{document}


\title{Arcane: An~\underline{A}ssertion~\underline{R}eduction Framework through Semantic~\underline{C}lustering~\underline{an}d MCTS-Guided Rule~\underline{E}xploring}

\author{
\IEEEauthorblockN{
Hongqin Lyu\textsuperscript{1,2},
Yonghao Wang\textsuperscript{1,2},
Zhiteng Chao\textsuperscript{1},
Tiancheng Wang\textsuperscript{1,2}
and
Huawei Li\textsuperscript{1,2}}
\IEEEauthorblockA{\textsuperscript{1}State Key Lab of Processors, Institute of Computing Technology, CAS, Beijing, China}
\IEEEauthorblockA{\textsuperscript{2}University of Chinese Academy of Sciences, Beijing, China}
}

\maketitle

\begin{abstract}
Assertion-based Verification (ABV) is essential for ensuring that hardware designs conform to their intended specifications. However, existing automated assertion-generation approaches, such as LLM-based frameworks, often generate large numbers of redundant assertions, which significantly degrade simulation efficiency. To mitigate the simulation overhead caused by redundant assertions, this paper proposes Arcane, an efficient assertion reduction framework. It integrates a two-tier assertion clustering approach for accurate semantic classification of large assertion sets, and employs Monte Carlo Tree Search (MCTS) to explore optimal rule-application sequences for efficient assertion reduction. The experimental results on Assertionbench \cite{b20} show that Arcane achieves a reduction of up to 76.2\% in the assertion count while fully preserving formal coverage and mutation-detection ability. Further simulation studies demonstrate a speedup of 2.6x to 6.1x speedup in simulation time. The proposed framework is released at~\url{https://anonymous.4open.science/r/Arcane1-0A6F/}
\end{abstract}

\vspace{-2pt}
\section{Introduction}
Functional verification is essential in integrated circuit (IC) design, where verification engineers check whether the designers' register-transfer level (RTL) code meets the architectural specifications. Assertion-based verification (ABV) is widely adopted in RTL design due to its ability to enhance observability and reduce simulation debugging time by up to 50\% \cite{b1}. In particular, high-quality SystemVerilog assertions (SVA) for property verification (PV) are critical within ABV methodologies \cite{b2}, particularly due to its ability to accurately reflect both high-level design intent and low-level RTL details. As the complexity of the design increases, the development of efficient methods for the generation of SVA has become an urgent necessity.


Traditional assertion generation approaches include schema-based methods that instantiate predefined templates \cite{b3}, as well as waveform- or trace-mining techniques such as HARM \cite{b4} and GoldMine \cite{b5}. While effective for certain pattern, these methods often produce semantically fragmented assertions and remain limited in handling complex design semantics. Hybrid rule-based and learning-based methods improve generation accuracy, but are still constrained by the complexity of natural language specifications \cite{b6}. More recently, large language models (LLMs) have opened new avenues for assertion generation \cite{b7}, leveraging strong semantic understanding to produce more comprehensive and context-aware assertions from specifications and RTL code while mitigating semantic fragmentation \cite{b8}. Researchers also explored LLM-based assertion generation in specialized scenarios such as hardware security \cite{b9}.

Although assertion generation techniques have advanced significantly, both traditional and LLM-based approaches still suffer from a key limitation: the generated assertions are often redundant and weakly integrated. Studies show that many assertions cover overlapping behaviors; for example, about 96\% of GoldMine’s assertions are redundant \cite{b10}. Similarly, LLM-based methods still produce 20\%–30\% redundant assertions. This redundancy increases simulation time and computational overhead, reducing verification efficiency, especially for large-scale designs \cite{b11}. Prior research has explored improving checking efficiency in both software and hardware domains, such as hybrid static-dynamic techniques for reducing runtime checking cost \cite{b12} and checker selection or refinement methods in hardware verification \cite{b13}. To the best of our knowledge, no prior work targets redundant assertion reduction while maintaining their error detection capability.




However, assertion reduction faces two major challenges. First, it is difficult to simplify assertions with complex boolean conditions and fine-grained temporal structures without changing their original verification intent, since direct or partial transformations may yield logically equivalent yet semantically confusing assertions that undermine the intended verification objectives. Second, the problem is inherently complex: as the number of assertions grows, interactions among logical conditions and signal usages increase exponentially, and the combinational use of reduction rules further enlarges the search space. 

To address these challenges, this paper presents Arcane, an efficient framework for assertion reduction. Arcane combines semantic embedding and automata-based acceptance analysis to capture behavioral similarities among assertions for high-quality clustering. It further adopts strictly semantics-preserving logical reduction rules and Monte Carlo Tree Search (MCTS) \cite{b14} to explore rule orders and combinations, maximizing redundancy elimination and reduction efficiency.

The contributions of this paper are summarized as follows:

\begin{figure*}[h]
\centering
\includegraphics[width=0.83\linewidth]{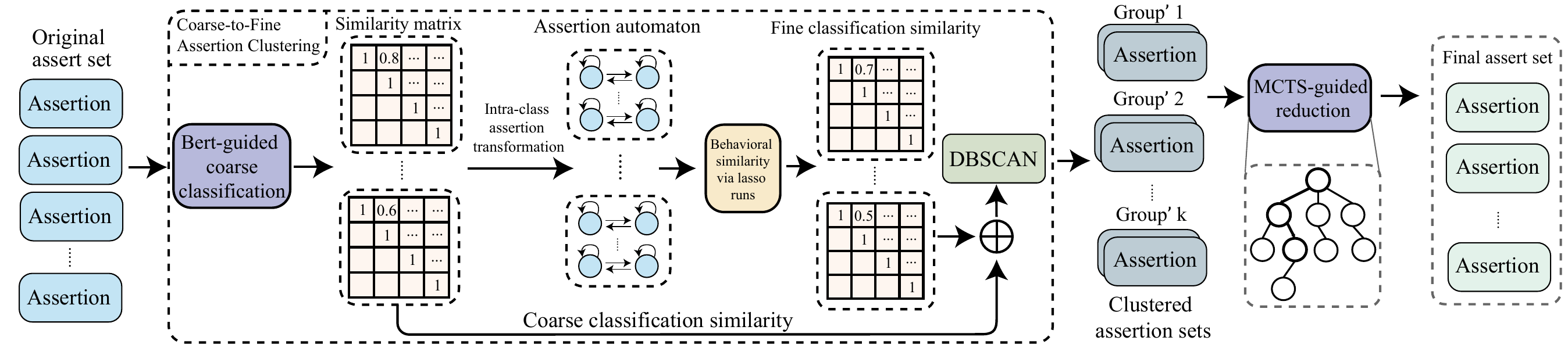}
\caption{The Arcane framework: It combines semantic and automata-based clustering with MCTS-guided reduction to generate a compact assertion set.
}
\label{fig:2}
\end{figure*}

\begin{enumerate}
\item A two-tier clustering scheme based on semantic embeddings and automata behavior is introduced to group assertions with consistent behavioral intent, avoiding interference during reduction.
\item Each reduction rule is mapped to an action in the MCTS search process. Arcane then uses MCTS to efficiently explore reduction paths within this action space, ensuring functional equivalence without changing the original verification constraints.
\item Experimental results show that Arcane preserves full verification quality while reducing assertion sets by up to 76.2\% and accelerating simulation by up to 6×. 
\end{enumerate}







\vspace{-2pt}
\section{Framework of Arcane}

Before diving into the design details, we introduce the overall flow of the proposed assertion reduction framework Arcane, which consists of two main stages: coarse-to-fine assertion clustering, and MCTS-based reduction, as illustrated in Figure \ref{fig:2}.

1$)$ \textbf{Coarse-to-Fine Assertion Clustering}: Assertions are translated into natural language, coarsely grouped by their similarity exported from the bidirectional encoder representations from transformers (BERT) \cite{b15}. The resulting groups are then refined with lasso-based fine-grained analysis and finalized by intra-class clustering.

2$)$ \textbf{MCTS-Guided Reduction}: For each refined cluster, MCTS is employed to search rule combinations that minimize redundancy while preserving coverage.

\subsection{Coarse-to-Fine Assertion Clustering}
\subsubsection{\textbf{BERT-guided coarse semantic classification}}
\mbox{}\\
To improve assertion reduction efficiency, assertions with similar functional intent are grouped together. Their formal and unambiguous semantics allow translation into equivalent natural-language descriptions, enabling meaningful classification. Such classification reduces the search space, and thus lowers the cost of fine-grained clustering through smaller, semantically consistent groups.


To achieve this, each assertion is first converted into a natural-language description. 
For example, the assertion $a \mathbin{\&} b \mathrel{\mid\!\to} c$ can be expressed as 
``if \textit{a} and \textit{b} hold, then \textit{c} must also hold in the same cycle.'' 
This aligns with BERT’s pretraining and improves semantic modeling. Each description is then encoded into a 1024-dimensional vector using a pretrained BERT model, and cosine similarity is used to group semantically related assertions for further fine-grained reduction.

\vspace{-2pt}
\subsubsection{\textbf{Behavioral similarity via lasso runs}}
\mbox{}\\
While BERT-based semantic classification enables coarse grouping, relying solely on natural-language similarity may cause misclassification, as textually similar assertions can represent conflicting functional conditions. For example, \texttt{assert} $(A \Rightarrow B)$ and \texttt{assert} $(A \Rightarrow \lnot B)$ are linguistically similar but functionally opposite. To address this, we introduce a lasso-based fine classification method that captures logic-level behavioral correlations among assertions.
\begin{figure}[ht]
  \centering
  \includegraphics[width=0.87\linewidth]{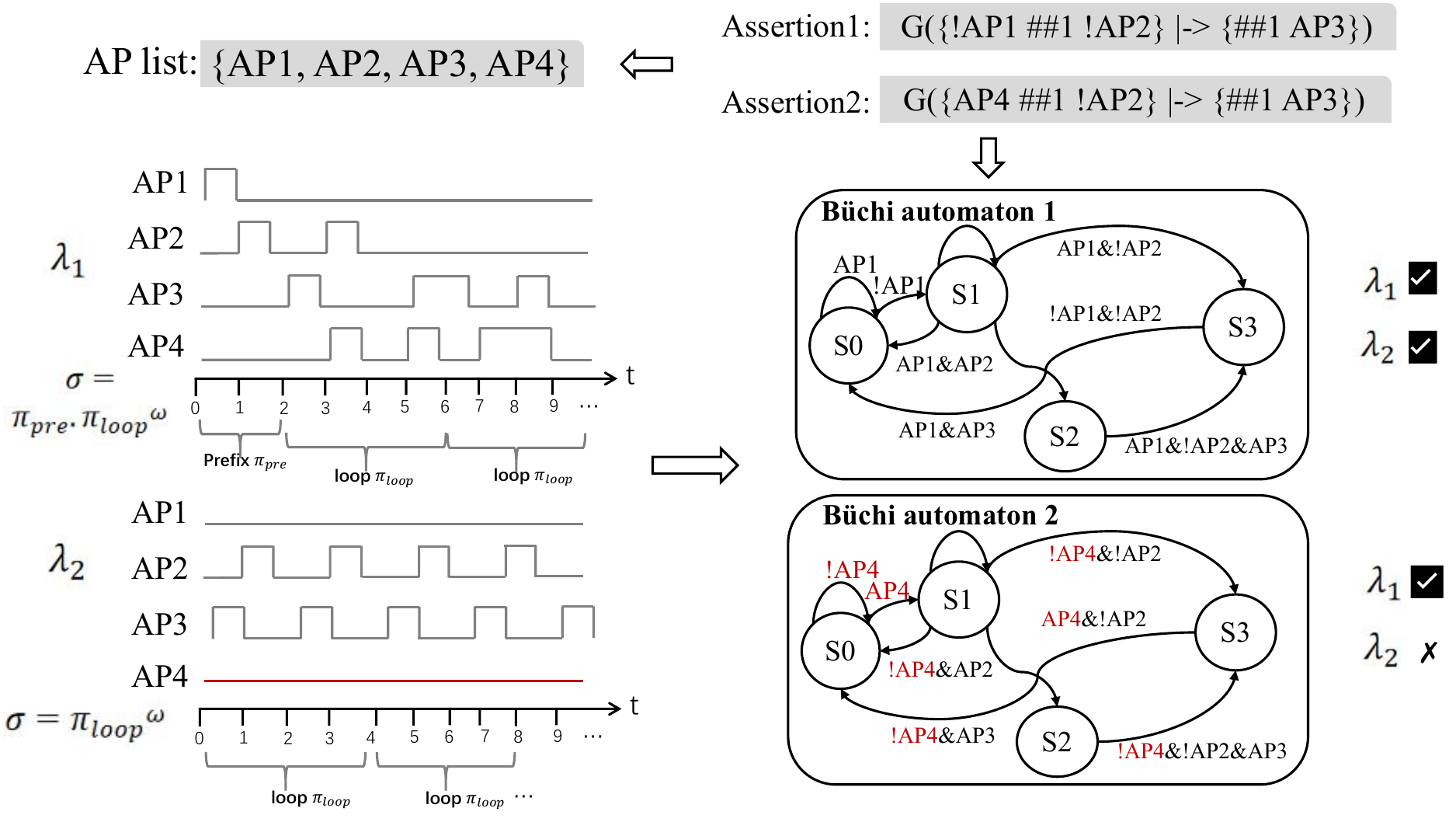}
  \caption{Both lasso paths $\lambda_1$ and $\lambda_2$ satisfy Assertion~1 and are accepted by B\"uchi automaton~1. B\"uchi automaton~2 requires AP4 to eventually be high, so it accepts $\lambda_1$ but rejects $\lambda_2$, whose loop keeps AP4 at 0.}
  \label{lasso}
\end{figure}

To capture assertion semantics beyond syntax, we analyze their acceptance behavior via lasso runs in Büchi automata \cite{b16}. A Büchi automaton is defined as 
\(\mathcal{A} = (Q, \Sigma, \delta, q_0, F)\), 
where \(Q\) is a finite set of states, \(\Sigma\) an input alphabet, 
\(\delta: Q \times \Sigma \rightarrow 2^{Q}\) the transition function, 
\(q_0\) the initial state, and \(F \subseteq Q\) the accepting set.
A \textit{lasso} represents an infinite execution by a finite prefix and a non-empty loop reachable from \(q_0\):
\begin{equation}
    \lambda = (\pi_{\text{pre}}, \pi_{\text{loop}}), \quad 
    q_0 \xrightarrow{\pi_{\text{pre}}} p \xrightarrow{\pi_{\text{loop}}} p,
\end{equation}
where \(p\) is the loop-entry state, \(\pi_{\text{pre}} \in Q^*\) is a finite prefix, and \(\pi_{\text{loop}} \in Q^+\) is a non-empty loop. 
The corresponding infinite trace is 
\(\rho_\lambda = \pi_{\text{pre}} \cdot \pi_{\text{loop}}^{\omega}\), 
where \(\cdot\) denotes concatenation and \(\omega\) denotes infinite repetition. 
The lasso \(\lambda\) is \textit{accepting} iff 
\begin{equation}
    \exists q \in F : q \in \pi_{\text{loop}},
\end{equation}
i.e., the loop visits at least one accepting state. Figure~\ref{lasso} illustrates this notion of lasso acceptance on two example traces and their corresponding Büchi automata.

After obtaining lasso representations, we quantify behavioral similarity based on acceptance over sampled runs. For each assertion \(a_i\), let \(\mathcal{L}_i = \{\lambda_1, \lambda_2, \dots, \lambda_m\}\) denote the sampled lassos, and \(\mathcal{A}_i \subseteq \mathcal{L}_i\) the accepted subset under \(\mathcal{A}_i\). The similarity between \(a_i\) and \(a_j\) is defined by the Jaccard index \cite{b17}:
\begin{equation}
    \text{Sim}(a_i, a_j) = 
    \frac{|\mathcal{A}_i \cap \mathcal{A}_j|}
         {|\mathcal{A}_i \cup \mathcal{A}_j|}.
\end{equation}
A higher value indicates more similar acceptance behaviors and thus stronger functional consistency.

To capture both linguistic and automaton-level behavioral similarity, we compute a unified score
$s_{ij}=\alpha\, s^{\text{NL}}_{ij}+\beta\, s^{\text{lasso}}_{ij}$ and convert it into a distance
$d_{ij}=1-s_{ij}$. We then apply DBSCAN~\cite{b18} to group assertions into dense semantic
clusters that form the basis for subsequent reduction.
\subsubsection{\textbf{Acceleration strategies for lasso-driven behavior classification}}
\mbox{}\\
The lasso-based characterization is expressive but computationally expensive. For a Büchi automaton $\mathcal{A} = (Q, \Sigma, \delta, q_0, F)$, the number of prefix–loop candidates is $O(|Q|^2 |\Sigma|)$, each requiring simulation for acceptance. To reduce cost, we first apply BERT-based classification to partition assertions into semantic subgraphs, and then perform lasso analysis locally, reducing computation while preserving behavioral fidelity.

For propositional assertions, we avoid lasso runs and instead enumerate all $2^N$ truth assignments over $N$ atomic propositions, compute satisfying sets, and use their Jaccard index as similarity. Since $N$ is typically small in hardware designs, this is more efficient than lasso-based processing.


\vspace{-2pt}
\subsection{MCTS-guided reduction}

After clustering assertions into semantically consistent groups, we perform intra-cluster reduction to eliminate redundancy. As different rule orders yield different results, we formulate this as a combinatorial optimization problem and apply MCTS to explore the search space. We model reduction as a deterministic MDP $(S, A, f, R)$, where states in $S$ represent assertion configurations, actions in $A$ apply reduction rules, $f(s,a)$ updates the set, and $R$ evaluates the resulting configuration, as detailed below.

\begin{figure}[ht]
  \centering
  \includegraphics[width=0.87\linewidth]{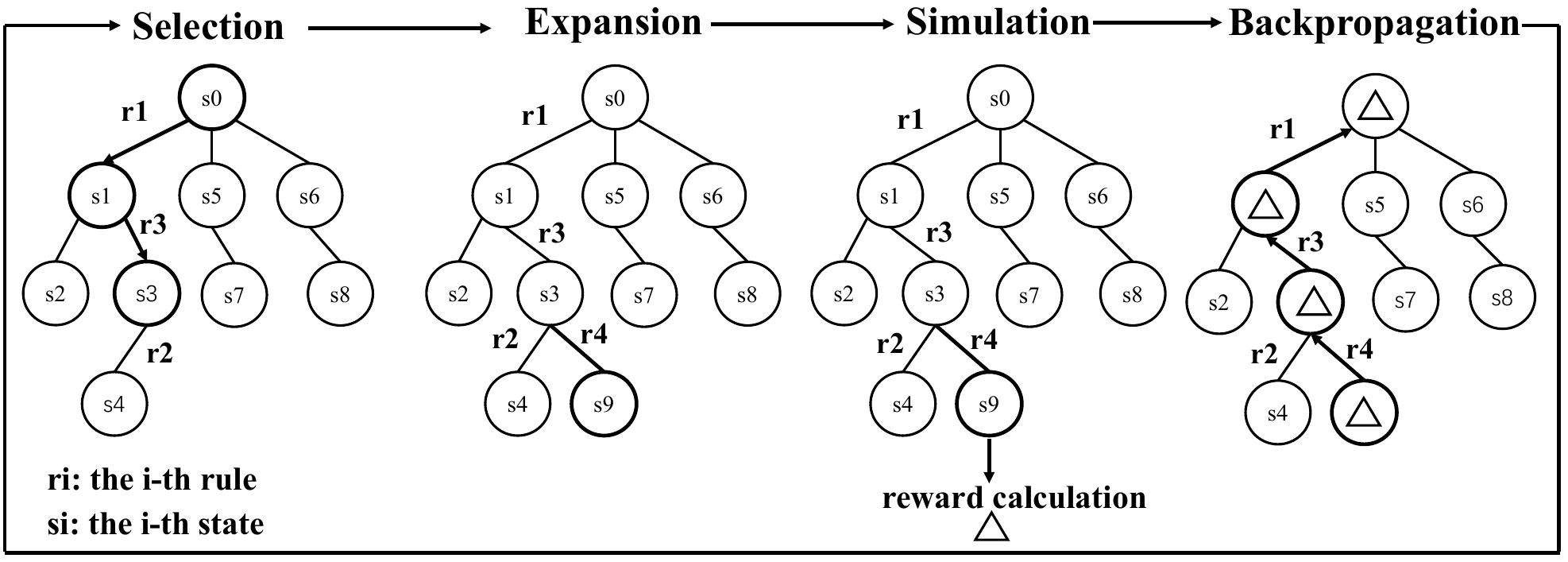}
  \caption{One iteration of the MCTS approach}
  \label{MCTS}
\end{figure}

Figure~\ref{MCTS} illustrates one iteration of MCTS. Starting from the root state $s_0$, the algorithm performs \textit{selection} by following tree edges to a leaf node, then \textit{expansion} by adding an unexplored child (e.g., $s_9$) via a rule $r_i$. A \textit{simulation} from this state yields a reward, which is \textit{backpropagated} to update statistics along the path. This four-step loop constitutes MCTS and is instantiated in our setting via the $S$, $A$, $f$, and $R$ components described below.

\textbf{State space.} In the context of assertion reduction, each state represents a specific configuration of the assertion set obtained after applying a sequence of reduction operations. Formally, a state $s_t$ corresponds to the assertion subset $S_t \subseteq S_0$, where $S_0$ is the original, unreduced assertion set.

\textbf{Action space.} The action space is composed of five predefined reduction rules, each representing a specific operation that modifies the current assertion set. The five rules are described below.
\begin{figure}[ht]
  \centering
  \includegraphics[width=0.9\linewidth]{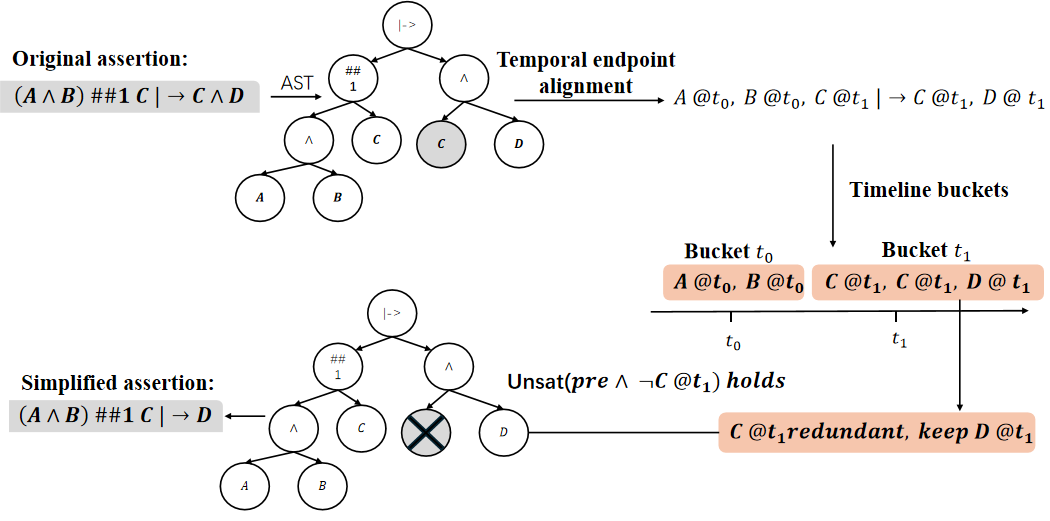}
  \caption{Example of intra-assertion reduction via temporal endpoint alignment and bucket-level entailment pruning.}
  \label{reex}
\end{figure}

\textit{Rule 1. Single-assertion pre/post reduction}:
At a high level, intra-assertion reduction applies sound Boolean simplification to time-normalized obligations. Temporal endpoint alignment rewrites operators into time-stamped propositions (e.g., \((A\ \#\#1\ B)\) becomes \(A @t_0\) and \(B @t_1\)), and timeline bucketing groups literals by time offset (e.g., \(t_0, t_1, \ldots\)), forming independent propositional obligations for local simplification.

As an example, consider a micro-rule within a single time bucket (Figure~\ref{reex}). 
For an obligation \(A: pre \mid\!\rightarrow post\) with 
\(post = \bigwedge_{\ell} \ell\), any literal \(\ell\) implied by \(pre\) can be removed without changing semantics:
\begin{equation}
post' = post \setminus \{\ell \mid pre \models \ell \}, 
\quad
A' : pre \mid\!\rightarrow post'.
\end{equation}
For instance,
\[
(A \land B)\ \#\#1\ C\ \mid\!\rightarrow\ C \land D
\;\;\Rightarrow\;\;
(A \land B)\ \#\#1\ C\ \mid\!\rightarrow\ D,
\]
since \(C\) is already implied by the antecedent. 
In each bucket, entailment \(pre \models \ell\) is checked via \(\mathsf{Unsat}(pre \land \neg \ell)\). 
This pruning only demonstrates one micro-rule; others are summarized in Table~\ref{Micro-rules}.

\begin{table}[t]
\centering
\caption{Micro-rules for non-temporal assertions.}
\label{Micro-rules}
\small
\setlength{\tabcolsep}{4pt}
\renewcommand{\arraystretch}{1.1}
\begin{tabular}{@{}p{0.48\linewidth} p{0.52\linewidth}@{}}
\toprule
\textbf{Micro-rule} & \textbf{Example} \\
\midrule
Reflexive removal 
& $P\ \mid\!\rightarrow\ P\ \Rightarrow\ \mathsf{del}$ \\[2pt]

Vacuous implication 
& $(A \land B)\ \mid\!\rightarrow\ A\ \Rightarrow\ \mathsf{del}$ \\[2pt]

Inconsistent antecedent 
& $(A \land \neg A)\ \mid\!\rightarrow\ C\ \Rightarrow\ \mathsf{del}$ \\[2pt]

Conjunct dropping 
& $A\ \mid\!\rightarrow\ (A \land C)\ \Rightarrow\ A\ \mid\!\rightarrow\ C$ \\[2pt]

Explicit falsity 
& $1\ \mid\!\rightarrow\ 0\ \Rightarrow\ \dots \mid\!\rightarrow 0$ \\[2pt]

Polarity conflict to zero 
& $A\ \mid\!\rightarrow\ (\neg A \land C)\ \Rightarrow\ \dots \mid\!\rightarrow 0$ \\[2pt]

Constant folding on consequent 
& $A\ \mid\!\rightarrow\ (C \land 1)\ \Rightarrow\ A\ \mid\!\rightarrow\ C$ \\[2pt]

Idempotence on consequent 
& $A\ \mid\!\rightarrow\ (C \land C)\ \Rightarrow\ A\ \mid\!\rightarrow\ C$ \\[2pt]

Absorption on consequent 
& $\dots \mid\!\rightarrow\ C \land (C \lor D)\ \Rightarrow\ \dots \mid\!\rightarrow C$ \\
Entailment pruning
& $(A \land C)\ \mid\!\rightarrow\ (C \land D)\ \Rightarrow\ (A \land C)\ \mid\!\rightarrow\ D$ \\
\bottomrule
\end{tabular}
\end{table}
\textit{Rule 2. Common-antecedent POST Conjunction}: When multiple assertions share the same clock annotation and antecedent, their consequents often capture distinct outcomes of the same triggering condition. Conjunctively merging these consequents (AND) reduces semantic redundancy while preserving full verification coverage. Mergeable assertions are identified by representing antecedents $P$ with their canonical AST-based keys $\kappa(P)$, which group syntactically different but logically equivalent forms together.

For each group, consequents are merged under the same timeline bucketing as in Rule~1, and a temporal rule is applied to handle SVA irregularities: eliminating redundant temporal prefixes (e.g., \texttt{\#\#1(\#\#1 C)} $\Rightarrow$ \texttt{\#\#2 C}).

\textit{Rule 3. Common-consequent PRE Disjunction}:
This rule is the dual of the conjunction rule, handling assertions with the same clock annotation and consequent. When multiple antecedents imply the same consequent, they are merged disjunctively to unify equivalent triggers and remove redundant paths:
\begin{equation}
\mathsf{Assert}\bigl(C\,(\!\bigvee_{i=1}^{m} P_i) \!\rightarrow Q\bigr),
\quad
\text{for } \{\mathsf{Assert}(C\,P_i \!\rightarrow Q)\}_{i=1}^m.
\end{equation}
For example,
\[
(A \!\rightarrow Q),\quad
(B \!\rightarrow Q)
\;\;\Rightarrow\;\;
\bigl((A \lor B) \!\rightarrow Q\bigr).
\]

For temporal cases, merging is performed within each timeline bucket, and identical delay structures are compacted into ranges (e.g., \texttt{\#\#[1:2]} $\parallel$ \texttt{\#\#[2:3]} $\Rightarrow$ \texttt{\#\#[1:3]}). Predicates with different structures or disjoint timing are not merged to preserve semantic equivalence.

\textit{Rule 4 \& 5. Pairwise Equivalence and Implication Determination}:
Logical relations between assertions are identified through pairwise analysis of their canonical Boolean forms. 
Given two reduced assertions \(E_1\) and \(E_2\), 
equivalence holds if \(E_1 \Leftrightarrow E_2\), and implication if \(E_1 \Rightarrow E_2\) or \(E_2 \Rightarrow E_1\). 
CNF-based entailment efficiently detects redundant or subsumed assertions, 
while inconclusive cases involving temporal dependencies are verified exactly using the \textit{SPOT} \cite{b19}model checker via Büchi automata comparison. 

\textbf{State transition.}
At each step, the application of a reduction rule updates the current assertion set, 
thereby producing a new configuration that reflects a different reduction stage. 
Formally, given the current state \(s_t\) represented by an assertion set \(S_t\), 
applying a rule \(r_i \in R\) yields a new state \(s_{t+1}\) with the updated set \(S_{t+1}\):
\begin{equation}
S_{t+1} = r_i(S_t).
\end{equation}
This transition reflects a concrete modification such as assertion merging, simplification, or elimination. 
Since each rule preserves semantic equivalence or containment, 
the overall reduction process forms a monotonically contracting sequence:
\begin{equation}
|S_{t+1}| \le |S_t|, \qquad
\llbracket S_{t+1} \rrbracket \subseteq \llbracket S_t \rrbracket.
\end{equation}
Thus, the reduction procedure evolves through a discrete sequence of states, 
each corresponding to a simplified assertion configuration.

\textbf{Reward mechanism.}
Each rule application yields a reward reflecting its reduction effect:
\begin{equation}
R_t = \Delta |S_t| + \Delta |AP_t|,
\end{equation}
where \(\Delta |S_t|\) and \(\Delta |AP_t|\) represents the decrease of assertion count and atomic predicate count respectively. 
This encourages transformations producing more compact yet semantically equivalent assertion sets.

Having defined the MDP formulation, the reduction process is optimized via MCTS, which explores rule application sequences to maximize the cumulative reward.

\textbf{MCTS optimization.}
The search adopts the Upper Confidence Bound for Trees (UCT) policy:
\begin{equation}
a^* = \arg\max_a \!\left[ Q(s,a) + c \sqrt{\tfrac{\ln N(s)}{N(s,a)}} \right],
\end{equation}
where \(Q(s,a)\) is the mean reward, 
\(N(s)\) and \(N(s,a)\) are visit counts, 
and \(c\) balances exploration and exploitation.
To accelerate convergence, the root is initialized with Rule~1 to encourage early variable reduction, and the search stops early if the maximal reward does not improve for three consecutive iterations. MCTS then expands promising nodes and selects the path with maximal cumulative reward for efficient semantics-preserving reduction without exhaustive search (Algorithm~\ref{alg:reduction}).

\begin{algorithm}[t]    
\setlength{\tabcolsep}{5pt}  
  \caption{Assertion Reduction with MCTS.}
  \label{alg:reduction}
  \begin{algorithmic}[1]
    \State \textbf{Input:} Clustered assertion groups $\mathcal{G} = \{G_1,\dots,G_K\}$, MDP $(S,A,f,R)$, reduction rules $\mathcal{R}=\{r_1,\dots,r_5\}$
    \State \textbf{Output:} Reduced assertion set $S^\star$

    \State $S^\star \gets \emptyset$

    \ForAll{$G_k \in \mathcal{G}$ \textbf{ in parallel}}
      \State $G_k^\star \gets \textsc{MCTS-Reduce}(G_k, S, A, f, R, \mathcal{R})$
      \State $S^\star \gets S^\star \cup G_k^\star$
    \EndFor

    \State \Return $S^\star$

    \Function{MCTS-Reduce}{$G_k, S, A, f, R, \mathcal{R}$}
      \State $s_0 \gets G_k$ \Comment{initial MDP state}
      \State root node $v_0$ with state $s_0$ and prior on $r_1$
      \State $R_{\max} \gets -\infty$, $c \gets 0$
      \While{$c < 3$}
        \State $(\{s_t\}, \{a_t\}, r) \gets \textsc{OneStepMCTS}(v_0, S, A, f, R, \mathcal{R})$
        \If{$r > R_{\max}$}
          \State $R_{\max} \gets r$, $c \gets 0$
          \State $G_k^\star \gets \textsc{ExtractAssertions}(\{s_t\})$
        \Else
          \State $c \gets c + 1$
        \EndIf
      \EndWhile
      \State \Return $G_k^\star$
    \EndFunction
  \end{algorithmic}
\end{algorithm}

\vspace{-2pt}
\section{Experiments}

\subsection{Experimental Setup}
In this study, we use AssertionBench \cite{b20}, which contains 112 hardware designs, each with RTL code, waveform traces, and corresponding assertions. These assertions are generated by two sources, HARM and an LLM-based generator, providing a realistic and heterogeneous benchmark for assertion reduction. For correctness analysis, we first filter assertions convertible to LTL \cite{b21}, then use Cadence JasperGold (v21.12.002) for formal verification and Synopsys VCS (v2016.06) for simulation. All experiments are conducted on a server with an Intel(R) Xeon(R) Gold 6148 CPU at 2.40 GHz and 629 GB RAM.

\subsection{Evaluation Metrics and Parameters}

To ensure objectivity and mathematical rigor, the assertions obtained after reduction are used directly without post-processing, and semantic equivalence is assessed using both Proof Core (PC) and Mutation Testing (MT, i.e., error-detection rate ER). Unlike structural metrics such as PC, MT more directly reflects assertion quality through mutation-detection effectiveness. A truly semantics-preserving reduction should therefore preserve both formal behavior and defect-detection capability. To quantify performance gains, we also measure reduction processing time (PT) and use VCS to evaluate simulation runtime (RT). A complete summary of the evaluation metrics is given in Table~\ref{Summary of Evaluation Metrics}.

\begingroup
\setlength{\abovecaptionskip}{3pt}
\setlength{\belowcaptionskip}{3pt}
\renewcommand{\arraystretch}{0.95}
\begin{table}[h]
\small       
\setlength{\tabcolsep}{5pt}  
\centering
\renewcommand{\arraystretch}{1} 
\caption{Summary of evaluation metrics.}
\label{Summary of Evaluation Metrics}
\begin{tabular}{c|c}
\hline
\hline
\cellcolor{gray!20}\textbf{Evaluation Metrics} & \cellcolor{gray!20}\textbf{Summary} \\ 
\hline
\makecell{$N$} & \makecell{\hspace*{-44pt} The number of FPV-passed SVAs}\\ 
\makecell{\textbf{$PC$}} & \makecell{\hspace*{-50pt} Minimal logic to prove property} \\
\makecell{\textbf{$ER$}} & 
\makecell{\hspace*{-28pt} Error detection rate in mutation testing} \\
\makecell{\textbf{$PT$}} & 
\makecell{\hspace*{-5pt} The processing time of the assertion reduction} \\
\makecell{\textbf{$RT$}} & 
\makecell{\hspace*{-26pt} The running time of the VCS simulation} \\
\hline
\hline
\end{tabular}
\\
\vspace{0.1cm} 
\scriptsize 
\raggedright 
\end{table}
\endgroup

\definecolor{deepgreen}{RGB}{110, 251, 152}
\definecolor{lightgreen}{RGB}{230, 230, 230}

\begingroup
\setlength{\abovecaptionskip}{3pt}
\setlength{\belowcaptionskip}{3pt}
\renewcommand{\arraystretch}{0.95} 
\begin{table*}[h]
\centering

\caption{Comparison of evaluation metrics before and after reduction.}
\setlength{\tabcolsep}{6pt}
\label{Combined Table}
\begin{tabular}{lccccccccccc}
\toprule
\multirow{2}{*}{\textbf{Design}} & 
\multicolumn{3}{c}{\textbf{$N$}} &
\multicolumn{2}{c}{\textbf{$PC$}} &
\multicolumn{2}{c}{\textbf{$ER$}} &
\multirow{2}{*}{\textbf{$PT(s)$}} &
\multicolumn{3}{c}{\textbf{$RT(s)$}} \\
\cmidrule(lr){2-4} \cmidrule(lr){5-6} \cmidrule(lr){7-8} \cmidrule(lr){10-12}
 & Orig. & Arcane & Ratio 
 & Orig. & Arcane
 & Orig. & Arcane
 & 
 & Orig. & Arcane & Ratio \\
\midrule
ca\_prng            & 576  & \cellcolor{lightgreen}137 & \textbf{76.2\%} 
                    & 92.0\% & 92.0\% 
                    & 3.2\% & 3.2\%
                    & 45.3
                    & 1595.28 & \cellcolor{lightgreen}261.93 & \textbf{6.1x} \\             
control\_unit       & 1447 & \cellcolor{lightgreen}441 & \textbf{69.5\%} 
                    & 99.3\% & 99.3\%
                    & 11.6\% & 11.6\%
                    & 209
                    & 7695.04  & \cellcolor{lightgreen}2223.98  & \textbf{3.46x} \\
eth\_cop            & 265  & \cellcolor{lightgreen}67  & \textbf{74.5\%} 
                    & 39.3\% & 39.3\%
                    & 0.9\% & 0.9\%
                    & 25.34
                    & 203.56 & \cellcolor{lightgreen}59.54 & \textbf{3.42x} \\
eth\_receivecontrol & 558  & \cellcolor{lightgreen}166 & \textbf{70.3\%} 
                    & 37.6\% & 37.6\%
                    & 12.5\% & 12.5\%
                    & 62.13
                    & 176.29 & \cellcolor{lightgreen}62.19 & \textbf{2.83x} \\  
MAC\_rx\_ctrl       & 2434 & \cellcolor{lightgreen}774 & \textbf{68.2\%} 
                    & 53.1\% & 53.1\%
                    & 39.1\% & 39.1\%
                    & 321.6
                    & 1786.12 & \cellcolor{lightgreen}670.03 & \textbf{2.67x} \\
MAC\_tx\_Ctrl       & 7161 & \cellcolor{lightgreen}2008 & \textbf{71.9\%} 
                    & 78.9\% & 78.9\%
                    & 49.8\% & 49.8\%
                    & 1388.33
                    & 10553.5     & \cellcolor{lightgreen}2866.41     & \textbf{3.68x} \\
\bottomrule
\end{tabular}
\end{table*}
\endgroup

We configure Arcane with fixed hyperparameters: 500 lasso samples, similarity weights of $\alpha = 0.4$ (BERT) and $\beta = 0.6$ (lasso), a unified similarity threshold of 0.85 for both coarse- and fine-grained classification, and 64 threads for pairwise similarity computation.
\subsection{Evaluation of the Arcane}
\begin{figure}[t]
  \centering
  \includegraphics[width=0.9\linewidth]{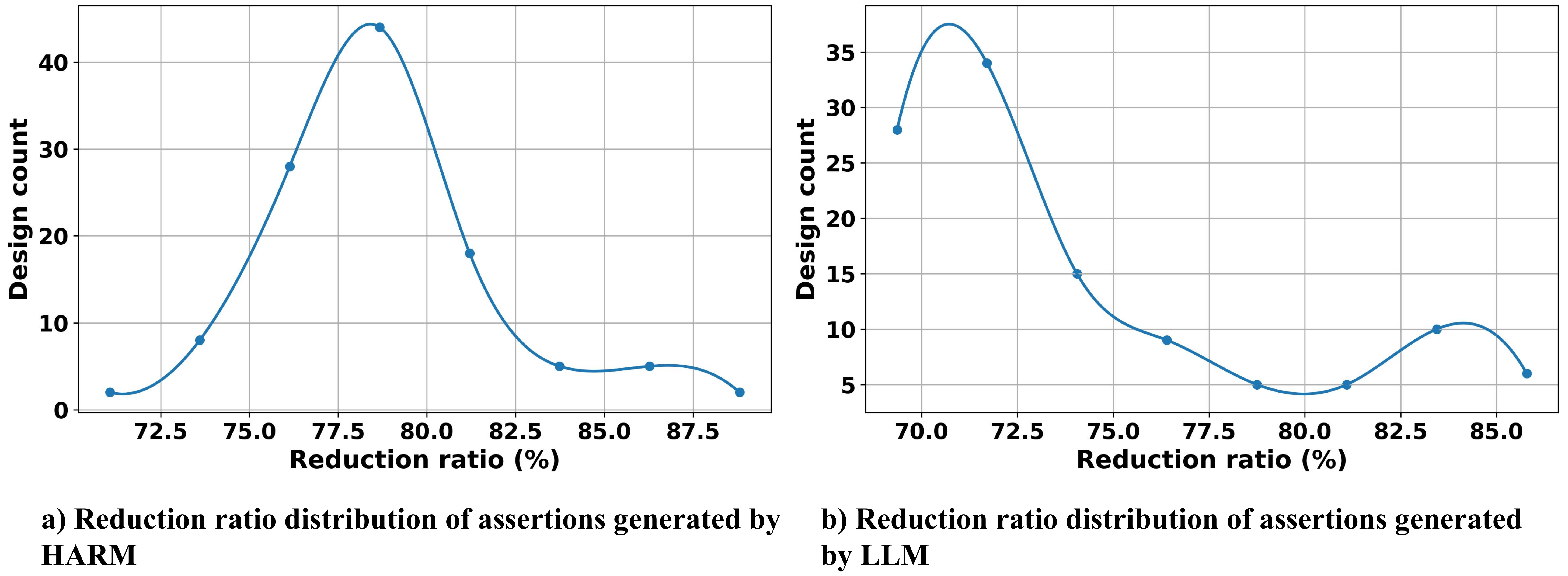}
  \caption{Overall distribution of reduction ratios achieved by Arcane across 112 benchmark designs.}
  \label{fig:reduction_heatmaps}
\end{figure}

\subsubsection{\textbf{Evaluation of assertion reduction ratio across full benchmarks}}

Figure\ref{fig:reduction_heatmaps} (a) and (b) illustrate the assertion reduction ratio distributions obtained by applying our framework to Harm‐generated and LLM‐generated assertions across the full benchmark set. Both curves are sharply peaked at \textbf{78 \%} and \textbf{71 \%}, respectively, confirming that assertion reduction for the vast majority of circuits converges to a high simplification level. The sharp drop-off in (a) shows that only a few cases deviate significantly, while (b) confirms that the assertion set for every design is reduced by at least \textbf{68 \%}, underscoring the stability of our framework.

\subsubsection{\textbf{Scatter plot distribution analysis of assertion clusters}}
\begin{figure}[ht]
  \centering
  \includegraphics[width=0.85\linewidth]{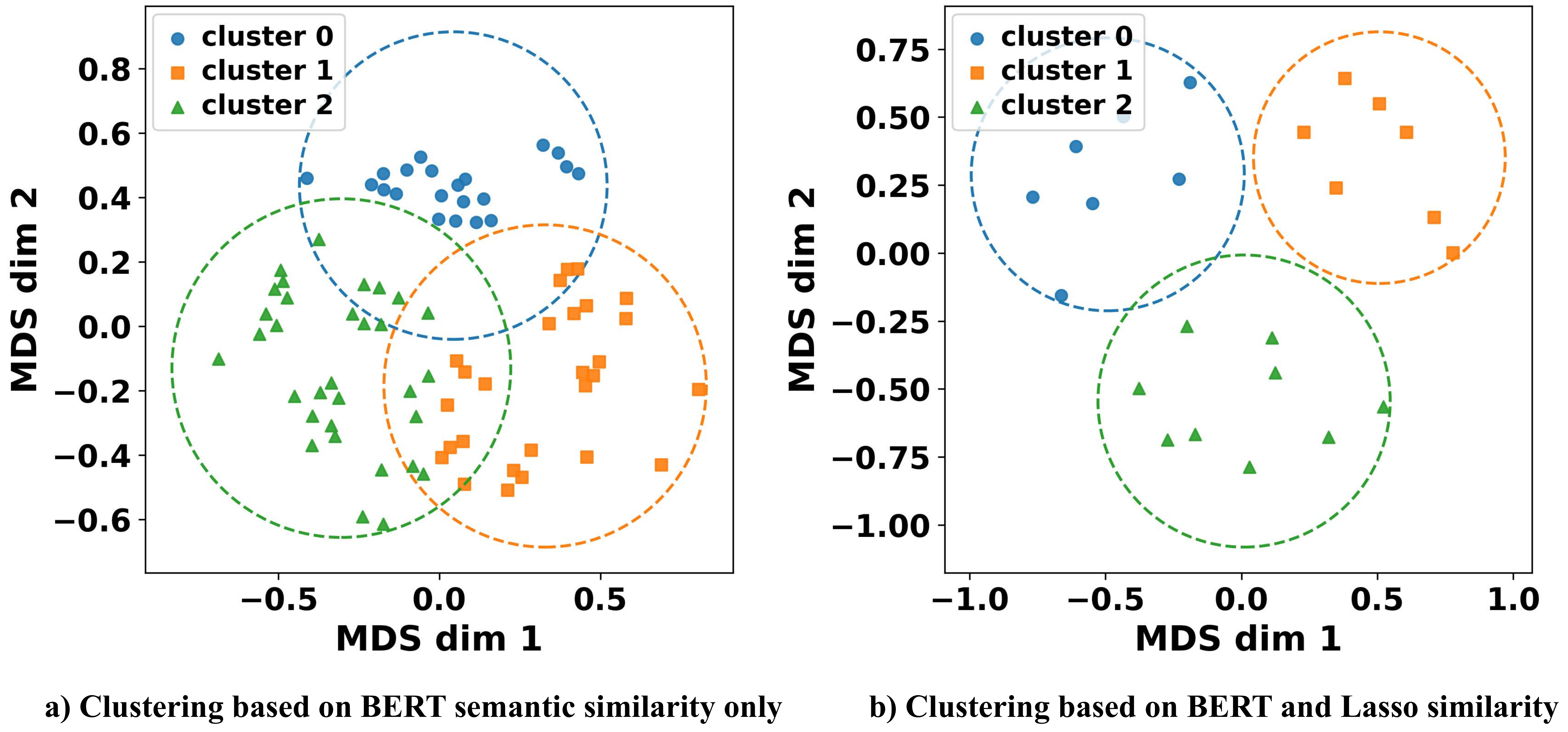}
  \caption{Scatter plot distribution analysis on assertion classification using BERT and BERT+LASSO similarity}
  \label{san}
\end{figure}

To visualize similarities among assertions, we construct a similarity matrix based on BERT-only similarity and BERT+Lasso similarity, and then apply MDS to embed it into 2-D space (Figure~\ref{san}). Using only BERT-based similarity, groups show overlap and loose structure, indicating limited ability to capture functional relations. With BERT+Lasso, groups become more compact with clearer boundaries, demonstrating improved intra-group cohesion and reduced inter-group coupling.


\subsubsection{\textbf{Coverage preservation and mutation detection}}

To ensure that assertion reduction does not degrade verification quality, we evaluate functional coverage and mutation-testing error-detection rates before and after redundancy elimination. All reduced assertion sets are executed on the same circuits under identical simulation configurations, and we select six relatively large benchmark designs to illustrate the results. As shown in Table~\ref{Combined Table}, Arcane reduces the number of assertions to roughly one quarter of the original size (\textbf{68\%–76\%} reduction) across all six designs, while both PC and ER remain exactly unchanged. These results demonstrate that the reduced assertion sets preserve full semantic equivalence to the originals in terms of formal proof coverage and defect-detection capability.

\begingroup
\setlength{\abovecaptionskip}{3pt}
\setlength{\belowcaptionskip}{3pt}
\renewcommand{\arraystretch}{0.95}
\begin{table}[t]
\centering
\small       
\setlength{\tabcolsep}{5pt}  
\setlength{\tabcolsep}{3.0pt}
\caption{Runtime and DBI of assertion classification with BERT-based semantic pre-classification.}
\label{tab:bert_runtime_dbi}
\begin{tabular}{lccc}
\toprule
\textbf{Design} & \textbf{Runtime(s) L$\rightarrow$B+L (×)} & \textbf{DBI (L)} & \textbf{DBI (B+L)} \\
\midrule
ca\_prng            & 144.45$\rightarrow$31.93 (\textbf{4.52})    & 0.3014 & 0.3172 \\
control\_unit       & 1070.55$\rightarrow$161.83 (\textbf{6.61})  & 0.3265 & 0.3379 \\
eth\_cop            & 30.14$\rightarrow$14.50 (\textbf{2.08})     & 0.2854 & 0.2907 \\
eth\_receivecontrol & 122.77$\rightarrow$51.06 (\textbf{2.40})    & 0.2937 & 0.3162 \\
MAC\_rx\_ctrl       & 3406.46$\rightarrow$289.51 (\textbf{11.77}) & 0.3374 & 0.3516 \\
MAC\_tx\_Ctrl       & 43321.48$\rightarrow$1274.51 (\textbf{33.99}) & 0.3419 & 0.3626 \\
\bottomrule
\end{tabular}
\end{table}
\endgroup

\subsubsection{\textbf{Acceleration performance before and after assertion reduction}}

To evaluate runtime benefits, we conduct VCS simulations with each benchmark running for 100k cycles. For the above six circuits, we apply our reduction framework to the original assertion sets and measure both reduction time and simulation time. Reduction time is a one-time cost, while simulation is repeated during verification and thus dominates runtime. As shown in Table~\ref{Combined Table}, our framework achieves an average simulation speedup of over \textbf{2.6$\times$}.

We also conduct an ablation study to evaluate the runtime impact of each component. Table~\ref{tab:bert_runtime_dbi} compares pure lasso-based classification (``L'') with BERT pre-classification followed by lasso (``B+L''). In B+L, assertions are first partitioned into semantically coherent classes, and lasso is applied locally, reducing clustering time from 30--3406s to 15--290s on mid-sized designs and from 43321s to 1275s on \texttt{MAC\_tx\_Ctrl} (\textbf{2.1$\times$--34.0$\times$}). The DBI (Davies-Bouldin Index, lower is better) \cite{b22} increase remains below 0.023 (0.015 on average), indicating only minor impact on clustering quality.
\vspace{-2pt}
\section{Conclusion}
This paper presents Arcane, a lightweight framework for eliminating redundancy in assertion sets. By combining BERT-based pre-classification, lasso-driven behavioral analysis, and MCTS exploration over a semantics-preserving rule space, Arcane achieves safe, equivalence-guaranteed reduction. Experiments on recent benchmarks show 68.2\%--76.2\% assertion reduction while preserving formal coverage and mutation detection capability, with 2.3--6.1$\times$ simulation speedup.
\vspace{-2pt}
\section{Acknowledgment}
This paper is supported by the National Natural Science Foundation of China (NSFC) under grant No. 92373206. The corresponding authors are Tiancheng Wang and Huawei Li.








\begin{thebibliography}{00}
\bibitem{b1} Y. Abarbanel, I. Beer, L. Gluhovsky, et al., ``Focs–automatic generation of simulation checkers from formal specifications,'' \textit{International Conference on Computer Aided Verification}, pp.538-542, 2000.
\bibitem{b2} H. Witharana, Y. Lyu, S. Charles, et al., ``A survey on assertion-based hardware verification,'' \textit{ACM Computing Surveys. (CSUR)}, vol. 54, no. 11, pp. 1-33, 2022.
\bibitem{b3} A. Hekmatpour, A. Salehi, ``Block-based shema-driven assertion generation for functional verification,'' \textit{IEEE 14th Asian Test Symposium. (ATS)}, pp. 34-39, 2005.
\bibitem{b4} Germiniani, Samuele, and Graziano Pravadelli. "Harm: a hint-based assertion miner." IEEE Transactions on Computer-Aided Design of Integrated Circuits and Systems 41.11, pp. 4277-4288, 2022.
\bibitem{b5} Vasudevan, Shobha, et al. "Goldmine: Automatic assertion generation using data mining and static analysis." 2010 Design, Automation \& Test in Europe Conference \& Exhibition (DATE), 2010.
\bibitem{b6} G. Parthasarathy, S. Nanda, P. Choudhary, et al., ``SpecToSVA: Circuit specification document to systemverilog assertion translation,'' \textit{Second Document Intelligence Workshop at KDD}, 2021.
\bibitem{b7} R. Kande, H. Pearce, B. Tan, et al., ``(Security) assertions by large language models,'' \textit{IEEE Trans. on Information Forensics and Security}, vol. 19, pp. 4374-4389, 2024.
\bibitem{b8} V. Pulavarthi, D. Nandal, S. Dan, et al.,  ``Are LLMs Ready for Practical Adoption for Assertion Generation?,'' \textit{IEEE Design, Automation \& Test in Europe Conference. (DATE)}, pp. 1-7, 2025.
\bibitem{b9} F. Aditi, M. S. Hsiao, ``Hybrid rule-based and machine learning system for assertion generation from natural language specifications,'' \textit{IEEE 31st Asian Test Symposium. (ATS)}, pp. 126-131, 2022.
\bibitem{b10} F. Wu, E. Pan, R. Kande, et al., ``Spec2Assertion: Automatic pre-RTL assertion generation using large language models with progressive regularization,'' \textit{arXiv preprint arXiv:2505.07995}, 2025.
\bibitem{b11} M. R. H. Iman, G. Jervan, T. Ghasempouri,  ``ARTmine: Automatic association rule mining with temporal behavior for hardware verification,'' \textit{IEEE Design, Automation \& Test in Europe Conference \& Exhibition. (DATE)}, pp. 1-6, 2024.
\bibitem{b12} Stulova, Nataliia, José F. Morales, and Manuel V. Hermenegildo. "Reducing the overhead of assertion run-time checks via static analysis." \textit{Proceedings of the 18th International Symposium on Principles and Practice of Declarative Programming.}, 2016.
\bibitem{b13} Sanjaya, Sahan, Hasini Witharana, and Prabhat Mishra. "Assertion-Based Validation using Clustering and Dynamic Refinement of Hardware Checkers." \textit{ACM Transactions on Design Automation of Electronic Systems 29.6}, pp. 1-22, 2024.
\bibitem{b14} Świechowski, Maciej, et al. "Monte Carlo tree search: A review of recent modifications and applications." Artificial Intelligence Review 56.3, pp. 2497-2562, 2023.
\bibitem{b15} Lee, J. D. M. C. K., and K. Toutanova. "Pre-training of deep bidirectional transformers for language understanding." arXiv preprint arXiv:1810.04805 3.8, pp. 4171-4186, 2018.
\bibitem{b16} Schewe, Sven. "Tighter bounds for the determinisation of Büchi automata." International Conference on Foundations of Software Science and Computational Structures. Berlin, Heidelberg: Springer Berlin Heidelberg, 2009.
\bibitem{b17} Niwattanakul, Suphakit, et al. "Using of Jaccard coefficient for keywords similarity." Proceedings of the international multiconference of engineers and computer scientists. Vol. 1. No. 6. 2013.
\bibitem{b18} Deng, Dingsheng. "DBSCAN clustering algorithm based on density." 2020 7th international forum on electrical engineering and automation (IFEEA), 2020.
\bibitem{b19} Duret-Lutz, Alexandre, et al. "Spot: a platform for LTL and omega-automata manipulation." \textit{Online: https://spot. lre. epita. fr}, 2023.
\bibitem{b20} Pulavarthi, Vaishnavi, et al. "Assertionbench: A benchmark to evaluate large-language models for assertion generation." Findings of the Association for Computational Linguistics: NAACL, 2025.
\bibitem{b21} Armoni, Roy, Dana Fisman, and Naiyong Jin. "SVA and PSL local variables-a practical approach." International Conference on Computer Aided Verification. Berlin, Heidelberg: Springer Berlin Heidelberg, 2013.
\bibitem{b22} Xiao, J., Lu, J., \& Li, X. . Davies Bouldin Index based hierarchical initialization K-means. Intelligent Data Analysis, 21(6), pp. 1327-1338, 2017.

\end{thebibliography}
\end{document}